\documentclass{article}
\usepackage{PRIMEarxiv}
\usepackage[utf8]{inputenc}
\usepackage[T1]{fontenc}

\usepackage{amsmath,amsfonts,bm}

\def\eqref#1{equation~\ref{#1}}

\def\1{\bm{1}}

\DeclareMathAlphabet{\mathsfit}{\encodingdefault}{\sfdefault}{m}{sl}
\SetMathAlphabet{\mathsfit}{bold}{\encodingdefault}{\sfdefault}{bx}{n}

\usepackage[hyphens]{url}
\usepackage{booktabs}
\usepackage{amsfonts}
\usepackage{microtype}
\usepackage{graphicx}
\usepackage{multirow}
\usepackage{subcaption}
\usepackage{array}
\usepackage{makecell}
\usepackage[round]{natbib}
\usepackage[title]{appendix}
\usepackage{xcolor}
\usepackage{tcolorbox}
\tcbuselibrary{breakable}

\definecolor{block-gray}{gray}{0.95}
\definecolor{xlinkcolor}{cmyk}{1,0.6,0,0}
\usepackage[bookmarks=false,
     pdfnewwindow=true,
     colorlinks=true,
     linkcolor=xlinkcolor,
     citecolor=xlinkcolor,
     filecolor=xlinkcolor,
     urlcolor=xlinkcolor,
     breaklinks=true,
final=true]{hyperref}

\makeatletter
\g@addto@macro{\UrlBreaks}{\UrlOrds}
\makeatother

\graphicspath{{figures/}}

\title{AgentNLQ: A General-Purpose Agent for \\ Natural Language to SQL \\
\vspace{0.5em}
}

\author{
\textnormal{Olena Bogdanov,\ Yeunji Jung,\ Chandra Dhir,\ Pareekshitreddy Gaddam,\ Saurabh Jain,\ Lakshmi Tumati} \\
\text{Vijay Parthasarathy$^1$,\ Anup Shirgaonkar$^1$} \\[0.5em]
\textbf{JPMorganChase} \\[0.3em]
{\small $^1$Corresponding Authors. \texttt{\{vijay.parthasarathy, anup.shirgaonkar\}@jpmchase.com}}
}

\begin{document}
\maketitle
\let\thefootnote\relax\footnotetext{\textbf{Disclaimer:} \textit{This paper was prepared for informational purposes and is not a product of the Research Department of JPMorganChase. JPMorganChase makes no representation, warranty or undertaking whatsoever and disclaims all liability for the completeness, accuracy or reliability of the information contained herein. This document is not intended as investment research or investment advice, or a recommendation, offer or solicitation for the purchase or sale of any security, financial instrument, financial product or service, or to be used in any way for evaluating the merits of participating in any transaction, and shall not constitute a solicitation under any jurisdiction or to any person, if such solicitation under such jurisdiction or to such person would be unlawful.}}

\begin{abstract}
Natural language to SQL (NL2SQL) conversion is an important problem for researchers and enterprises due to the ubiquitous importance of relational databases in broad-ranging practical problems. Despite the rapid advancements in the capabilities of LLMs, NL2SQL has not reached parity in accuracy with human expert SQL writers, hence needing additional improvements in NL2SQL algorithms. This study presents a new multi-agent method for NL2SQL that achieves 78.1\% semantic accuracy on the BIg Bench for LaRge-scale Database (BIRD) benchmark. Our method leverages a semantically enriched representation of user-provided schema, adds user-provided business rules, and produces accurate SQL queries. The main contributions of this study are (a) We designed an optimized new orchestrator in a multi-agent solution that uses LLMs to plan, orchestrate, reflect, and self-correct to generate accurate SQL queries, (b) We developed an advanced schema enrichment method that creates context-aware metadata to improve accuracy, and (c) We demonstrated the accuracy and generalizability of the method across different domains and datasets by evaluating it on the BIRD-SQL benchmark.
\end{abstract}
\section{Introduction}

Relational databases store large quantities of business data in nearly every enterprise, making efficient access to this information a critical challenge. Extracting insights from structured databases often requires writing SQL queries, a specialized technical skill that may limit business users from obtaining quick answers and insights from their data. Technically skilled data practitioners also spend a significant portion of their time finding, cleaning, and organizing data (including writing and debugging SQL). Therefore, automatic generation of SQL queries from natural language questions is a highly desirable capability for modern enterprises.

Recent developments in Large Language Models (LLMs) have spurred extensive research into automating this task. AI-enabled tools have automated SQL query authoring to a large extent \citep{mohammadjafari2024}. With these tools, users can ask natural-language questions about their databases, and an LLM generates the corresponding SQL query and can execute it on the user's behalf. Typically, single-instance LLM calls or general-purpose question-answering LLMs are limited in their performance on SQL generation, as they do not have knowledge of the database schema, business rules, and they often cannot reliably perform multi-step reasoning or verify and self-correct using execution feedback.

Enterprise-grade SQL understanding demands an even deeper knowledge of databases within a business unit: table structures, column names, relationships, and business definitions. Analysts are often tasked with searching through Entity-Relationship diagrams, learning domain-specific constructs (e.g.\ how ``customer'' is labeled in a particular schema or what the difference between ``transaction type'' is across tables in a database) which takes a significant amount of SQL-writing time. While ``gold-standard'' SQL writers can achieve very high accuracy rates (e.g.\ 90\% or more on BIRD-Bench), in real-world scenarios factors such as schema complexity, evolving requirements, complex joins, and ambiguous requests can lead to lower accuracy compared to optimized benchmark environments. As a result, generating accurate SQL queries requires a large amount of manual authoring or oversight, even when an LLM is used to generate queries.

Improved LLM and agent reasoning has enabled the development of more generalizable tools, reducing the need for ad-hoc, task-specific solutions which traditionally constrained engineering teams. Multi-agent solutions are known to enhance the reasoning capability of LLM-based solutions and provide an effective means to improve the ability to generate complex SQL queries \citep{googlecloud2025}.


We present AgentNLQ, an agentic solution that combines several elements: (1) multi-agent orchestration with self-reflection and Chain-of-Thought (CoT) reasoning for iterative refinement, (2) automated metadata generation that enriches database schemas for improved database understanding, and (3) a multi-model agent configuration that leverages the complementary strengths of different LLMs. AgentNLQ unifies these capabilities into a single end to end solution for generating SQL queries from natural language questions.

\section{Related Work}

NL2SQL tasks involve many technical and scientific challenges, including lexical and syntactic ambiguity in users' questions, recognizing entities and temporal concepts (e.g.,\ ``last year''), output format (free-form vs. structured output vs. data frame), multiple valid SQL queries for a single correct answer, and the need to understand complex relationships among a large number of columns across multiple tables \citep{liu2025survey}.

Recent work from public benchmark leaderboards such as BIRD \citep{birdsql2025} and Spider \citep{spider2025} shows some trends in approaches used to solve these challenges using LLMs, and various methods of optimizing such approaches. They include leveraging the reasoning ability of LLMs to plan and orchestrate SQL generation \citep{zhai2025excot}, schema linking to efficiently search for tables and columns most relevant to the user query \citep{deng2025reforce}, ensembles (majority voting) \citep{xiyansql2025}, reinforcement learning and finetuning \citep{cohere2025}, dynamic few-shot selection, and constructing synthetic few-shot examples to guide the LLM SQL generation \citep{pourreza2024chase}.

Among the commercial offerings, Snowflake's research combines reasoning of LLMs via Chain of Thought (CoT) with preference optimization \citep{zhai2025excot}. This presents an example of leveraging inference time scaling to improve accuracy. Inference-time scaling is also used in ensembles \citep{contextualai2025, xiyansql2025} and by generating a large number of candidate SQL queries which are then filtered by majority voting or a reward model \citep{wang2023macsql, contextualai2025}. The key finding from these studies is to base the reward model on execution of the SQL, as opposed to basing it on generated text feedback, which is a more common paradigm in preference learning for LLMs.

We find that multi-agent approaches have emerged as a promising method to improve accuracy. For example, \ MAC-SQL \citep{wang2023macsql} introduces an architecture with three separate agents for schema linking, query decomposition, and generation. Other studies also focus on agents for preprocessing and custom chain of thought pathways \citep{pourreza2024chase}. We take inspiration from research focused on generalist multi-agent systems for open-ended tasks and adapt the Magentic One Orchestrator Agent \citep{magneticone_blog} for our current version of the NL2SQL solution due to its detailed planning ability. We also designed a new custom orchestrator which is optimized for efficient planning and orchestration. This orchestrator is described later in Fig.~\ref{fig:orchestrator}.

In addition to the above techniques, prior literature shows that including few-shot examples (in-context learning) in the context for the SQL generation LLM improves accuracy, especially when precise format and syntax are required to produce the correct SQL query. Several leaderboard solutions use few shot examples in their mix of optimizations \citep{pourreza2024chase, xiyansql2025, contextualai2025, floratou2024nl2sql}. Contextual AI's method \citep{contextualai2025} uses just one example from their training set, which they found to improve accuracy. Therefore, we leverage few shot examples, when available, as inputs from the user.

Schema linking is also essential for producing highly accurate SQL. This process involves sophisticated reasoning to interpret column names, table names, and table data, enabling the identification of foreign keys, relationships between tables, and the semantic significance of columns and their values, all relevant to a specific use case. Recent studies, such as discussed earlier, leverage various adaptations of schema linking. This includes probing the database to explore these complex relationships, understanding and interpreting them, and encapsulating them with an enriched schema description that captures these links across many columns. Prior studies have adopted LLM-based methods to do this and have used exploratory data analysis type probing \citep{dragusin2025}.

Building on these findings, our approach incorporates several of these techniques. We allow users to provide few-shot examples relevant to their use cases and provide business rules in the context along with the schema. We constructed a new schema enrichment process, which leverages LLMs to write rich column and table descriptions, identify entities, and probe the database to map schema linkages across multiple tables.

\section{Methodology}

\subsection{Problem Formulation}

Given a natural language question and relevant business-specific knowledge about a database, the goal of NL2SQL is to generate an accurate and executable SQL query. This problem can be formalized as the following optimization problem:

\begin{equation}
\label{eq:formulation}
\text{SQL} = \underset{k \leq k_{\max}}{\operatorname{argmax}} \; E\!\left(\text{LLM}_{\theta}\!\left(Q_{\text{NL}},\, S,\, \lambda_{\text{biz}}\right),\, k_{E_{\max}}\right)
\end{equation}

\noindent where
\begin{itemize}
\item $E$ = Evaluation objective function that assesses the correctness of the generated SQL query
\item $Q_{\mathrm{NL}}$ = Natural language question from the user
\item $S$ = Schema of the SQL database
\item $\lambda_{\mathrm{biz}}$ = Business specific rules that include the use case's domain knowledge
\item $\mathrm{LLM}$ = SQL generator LLM, with $\theta$ as its hyperparameters
\item $k_{E_{\max}}$ = Number of self-reflective attempts done by the agent to arrive at the best query (early stopping)
\item $k_{\max}$ = Maximum attempts allowed for the agent to produce its best possible SQL query
\end{itemize}

\subsection{Agentic Approach}

We developed a multi-agent orchestrated workflow with a planner / orchestrator, SQL generator, and query execution agents with self-learning and delegation abilities. This is paired with a one-time rich schema representation generation. Figure~\ref{fig:architecture} shows the user-provided inputs, and the agent's internal working methodology.

The agent works in two stages: metadata generation and inference. First, an offline pipeline generates metadata and a vector index from the user's database schema, for use at inference time. The metadata contains several components: (1) a statistical description of the data, (2) primary and foreign key mappings, and (3) LLM-generated table and column descriptions. The metadata generation step retrieves primary and foreign keys, or derives them if unavailable. An LLM then combines the user-provided schema and this extracted knowledge to generate descriptions of each table, column, and their relationships within the database. This metadata is then embedded and searched at inference time to supply additional context to our agents (Fig.~\ref{fig:architecture}).

The second stage is inference, where we begin by extracting relevant entities from the user's question. These entities -- keywords that may be relevant to tables, columns or values -- are embedded and retrieved using vector search. The metadata for the relevant tables and columns is then retrieved and injected into the system prompt of our Orchestrator and SQL Generator agents. We also introduce an optimization that checks the total schema token length at runtime. If the token length is below the model's context limit, we include all tables and columns directly, bypassing vector search to reduce latency. We also incorporate any additional user-provided information to ground the agent in the user's business domain, including relevant business rules and few-shot examples.

To assist the agent's chain of thought, we provide a `scratchpad' (referred to as the ``fact sheet'') that decomposes SQL generation into several steps, including query decomposition, identifying relevant filters and measures, join path identification, global or local aggregations, WHERE clauses, and column existence checks. We developed an NL2SQL orchestrator agent that employs a reason-generate-evaluate-replan chain of thought with two distinct ledgers: a fast-thinking System~1 loop for standard execution, and a slow-thinking System~2 loop for error recovery.

\begin{figure}[t]
\centering
\includegraphics[width=0.8\textwidth]{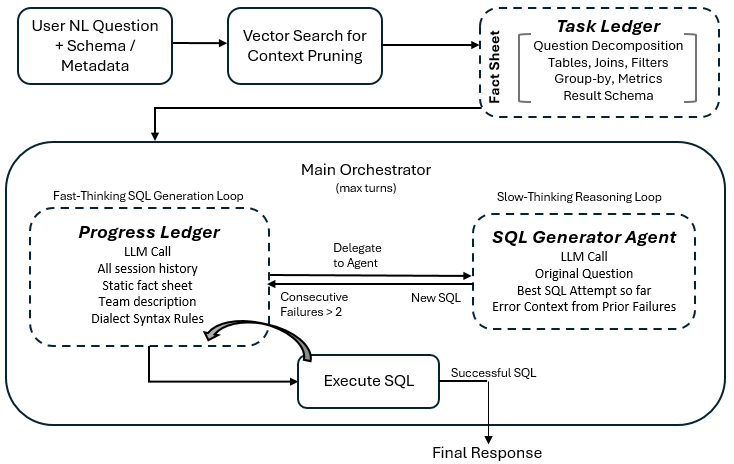}
\caption{The NL2SQL orchestrator is a state machine that tracks the overall state of task progress and dynamically selects execution paths through different states of goal completion. The main orchestrator serves as both the System~1 (fast) pathway -- emitting and validating SQL directly via syntactic parsing and execution -- and the escalation controller that delegates to the System~2 (slow) SQL generator agent for reasoning-intensive query construction when the fast path fails.}
\label{fig:orchestrator}
\end{figure}

The orchestrator attempts to solve the task by generating successive iterations of SQL queries, invoking the SQL execution tool each time to obtain direct execution feedback. The feedback includes the SQL result upon successful execution, or an error message and failure reason upon failed execution. The orchestrator also has the ability to self-reflect on the accuracy of the generated SQL for the user's question given the schema information. The orchestrator can decide successful task completion, or after three unsuccessful generations, delegate control to an SQL author agent which rethinks the entire plan, adjusts the approach if necessary, and runs a detailed plan-author-execute-reflect loop. Additional details are provided in the orchestrator section below (Fig.~\ref{fig:orchestrator}).

\subsection{Key Components}

\subsubsection{Models}

We use OpenAI's text-embedding-3-large-1 as our embedding model. For the agent group, we use OpenAI GPT-4o or Anthropic Claude Opus 4.1 with temperature 0, assigning the LLM best suited for each task (planning and SQL generation). We implement an orchestrator agent, query generator agent, and a query executor tool. We experimented with Autogen's Magnetic-One orchestrator \citep{magneticone_blog, fourney2024magneticone}, as well as our new custom NL2SQL orchestrator. We present results from both (Table~\ref{tab:accuracy_domains}), and show that our custom orchestrator achieves better accuracy and lower latency than the Autogen orchestrator.

\begin{figure}[t]
\centering
\includegraphics[width=\textwidth]{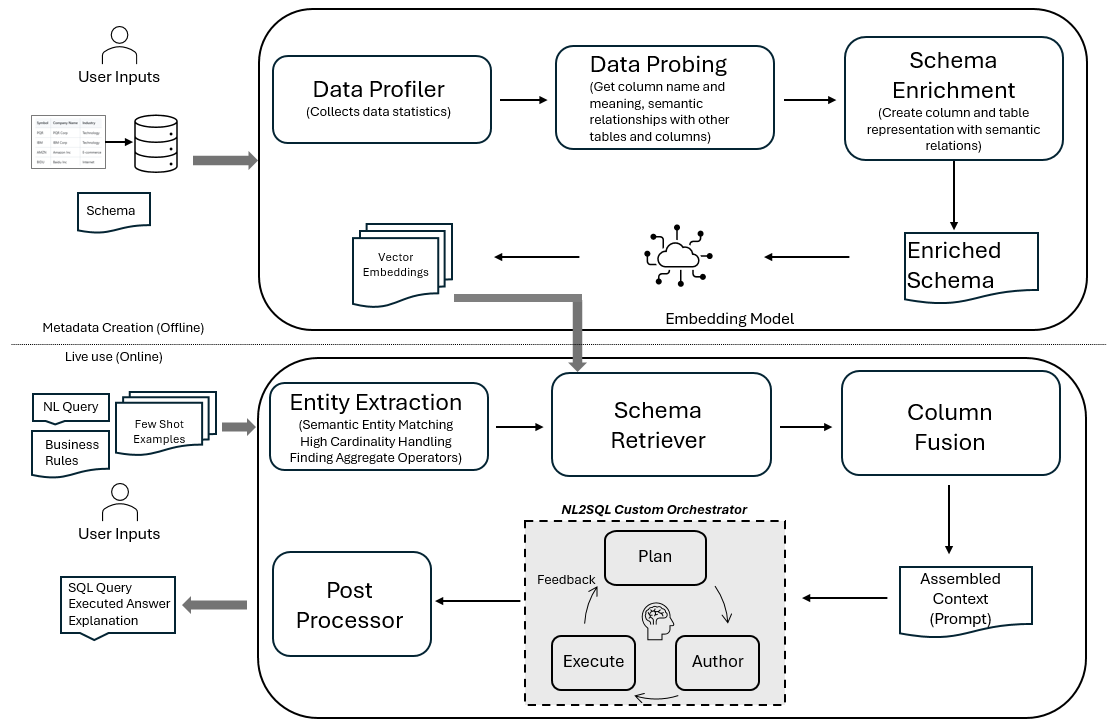}
\caption{NL2SQL agent architecture. An offline metadata generation pipeline probes the SQL database and schema and prepares an enriched metadata. At serving time, the orchestrator plans and executes a reasoning-based plan with a multi-agent setup with self-critic and iterative refinement via detailed chain of thought.}
\label{fig:architecture}
\end{figure}

\subsubsection{Vector Search}

When the enriched schema token length is below the model's supported context limit, we use the entire enriched schema in the query generation context to provide maximum information to the query planner and writer agents. However, when the metadata is large (exceeding the token limit of the LLM), pruning the context in a naive way risks losing useful information. Therefore, we perform vector search over all (table, column) combinations in the entire database, and retrieve the top-$k$ columns that best match the entities from the user's query. This provides the agents with the required columns that are essential to answer the user's question, thereby improving accuracy. During our experiments, the majority of datasets in BIRD were able to ingest the entire enriched schema. However, when the context gets large, vector search is needed to maintain accuracy or even to avoid failure due to context length overflow. (See Table~\ref{tab:ablation} for ablation study showing vector search vs. no vector search accuracy.)

\subsubsection{Metadata Generation}

NL2SQL benchmarks such as SPIDER \citep{lei2025} and BIRD contain extensive metadata that is generally not available in practice. Human-generation of metadata requires both subject matter expertise and knowledge of the underlying data schemas, making it time consuming and expensive. We developed a metadata preparation algorithm to automate this without requiring any human intervention. In this preprocessing step, we aggregate characteristics about our database, including the number of null, distinct, and example values, primary and foreign keys. Further, we use an LLM to generate an information dense description of our data that combines key descriptive characteristics from the previous metadata generation step (data description), and user supplied domain knowledge (schema and business rules). Our findings show that using metadata improves NL2SQL accuracy, even after removing examples supplied in the prompt at the inference step (Table~\ref{tab:ablation}).

\subsubsection{Prompt Engineering}


Prompt engineering in AgentNLQ centers on a schema-focused strategy: the inference pipeline assembles dynamic context (metadata slices, foreign key graph, extracted values, column mappings with aliases, examples, and validation checklists), and task-based prompting on join complexity checks, filter aggregation and verification. We re-prompt the agent with the question at the end of the prompt to ensure adherence to the central task.

\subsubsection{NL2SQL Orchestrator}

One of the main contributions of this work is a domain-specialized, ledger-driven orchestrator for NL2SQL. Dynamic goal progress tracking using ledgers has been used as an effective way to solve tasks with multiple agents working together. Magentic One \citep{fourney2024magneticone} has been a notable paradigm that employs a dual-ledger architecture (task ledger and progress ledger) designed for open-ended tasks such as web browsing, file system interaction, and general code generation. While effective in those settings, this generally introduces unnecessary overhead for SQL generation in a production setting, a targeted goal with stringent latency constraints and a well-defined success criteria (a single executable query that answers the user's question).

We adapted the dual-ledger paradigm specifically for NL2SQL by separating out the static part of the goal context into a task ledger, and the dynamic reasoning part into a live progress ledger. This retained the structured planning and iterative refinement strengths while eliminating the overhead of updating two separate ledgers. The key design elements are:
\begin{enumerate}
\item Collaboration of System~1 (fast) and System~2 (slow) thinking
\item Direct SQL validation and execution feedback
\item Efficient context management
\end{enumerate}

\subsubsection{Collaborating Fast/Slow Reasoning Systems}

We combined fast (System~1) and slow, reasoning intensive (System~2) loops that operates in tandem.

The fast thinking loop, which is the orchestrator's LLM itself, attempts to resolve the queries by using the compact fact sheet and pruned execution feedback, with a configurable number of self-improvement rounds. At each round, the orchestrator evaluates the current state and decides among three actions: (i) emit directly: generating or correcting a SQL query itself and immediately validating it via syntactic parsing (using SQLGlot) and execution against the database, (ii) delegate to the SQL generator agent for reasoning-intensive query construction, or (iii) delegate to the SQL executor tool for execution and result validation.

The orchestrator remains the central decision-maker throughout, with the context managed by the progress ledger described below. This enables the orchestrator to resolve straightforward queries in one or two rounds with minimal token expenditure. Only when it cannot produce a correct query within a configurable number of consecutive attempts (default: 2), the orchestrator escalates to the reasoning-intensive SQL generator, which forms the slow reasoning based system.

The slow thinking loop, on the other hand, is the separate SQL generator agent that has an elaborate prompt that activates detailed chain of thought reasoning. It receives the full metadata context along with session history of past failed attempts (see details in the section on efficient context management below). Its session memory acts as a scratchpad for detailed reasoning. It considers the whole context in entirety, and authors the next version of the query. Due to it having the full context, as well as reasoning chain, it is slower but more accurate, and serves as a fault recovery agent for the complex queries which the fast thinking loop could not solve. The SQL generator then returns the new generated query back to the orchestrator.

Upon receiving the generator's output, control returns to the orchestrator loop, which validates the result and either finalizes the answer or continues iterating. At this point it can self-determine to either emit this query after validation, or regenerate it itself, or again escalate to the slow thinking agent for further refinement. The orchestrator bounds the total worst-case retry budget to four attempts combined for both systems before returning a best-effort answer. This dual-process design allows the system to resolve the majority of queries cheaply via the fast pathway while reserving expensive, reasoning-intensive generation for cases that genuinely require it.

\subsubsection{Direct Execution and Validation for Feedback}

We use execution-grounded feedback via syntactic validation, live query execution, and result-set heuristics rather than abstract LLM self-assessment. Rather than relying on the LLM to assess progress through abstract reasoning, the progress ledger validates each candidate query through concrete signals: syntactic parsing via SQLGlot, execution against the database, and result-set heuristics (e.g.\ row-count magnitude checks, empty-result detection). This grounds every iteration in observable outcomes, reducing hallucinated self-assessments. This is similar to the direct feedback used by some previous studies \citep{zhai2025excot}.

\subsubsection{Efficient Context Management}
The orchestrator and its fast and slow loops are supported by two ledgers: a task ledger and a progress ledger. Unlike the Magentic One orchestrator which dynamically maintains both ledgers, we consolidate the dynamic context management into progress ledger and keep task ledger to only contain the immutable part of the goal plan.

The task ledger is invoked once at initialization to compile the immutable fact sheet. It receives the full database metadata and the user question, and produces a compact, structured JSON plan containing sub-questions with dependency ordering, required tables, join paths, filters, group-by clauses, metrics, and business rules. This artifact serves as the persistent goal representation throughout the orchestration lifetime. This compaction distills the full metadata into a compact, information-dense representation that is a fraction of the original token cost, enabling a fast-process orchestration pathway in which the progress ledger can reason over the plan and emit or correct SQL with minimal prompt overhead. In the fast loop, the task sheet only pairs with a selectively pruned session history that retains only high-signal execution feedback.

The progress ledger includes all of the dynamic context to support the System~2 loop. After the slow thinking system is activated in the orchestrator, the SQL generator receives a maximally enriched prompt -- comprising the full schema markdown, dialect-specific instructions, and a structured scratchpad that supports chain-of-thought reasoning -- thereby exploiting the model's context window to its limit. When available metadata exceeds this limit, a lightweight vector-similarity retrieval stage truncates the context by populating it with the most relevant schema elements in ranked order.

To prevent context degradation, the orchestrator actively manages the context in the progress ledger. During iterative refinement, the agent context is selectively filtered to include only execution-based feedback (e.g.\ error messages and result summaries), thereby preserving high information density and conserving tokens. To guard against context degradation over extended interaction chains, we employ a pruning strategy when the conversation history approaches the model's context limit: the immutable fact sheet is always retained, along with the feedback from the first and most recent attempts, preserving both the initial grounding signal and the latest error diagnostics. To mitigate the well-documented needle-in-a-haystack problem \citep{kamradt2026}, we re-append the user's original question at the end of the prompt, anchoring the generator's attention to the stated goal. Together, these mechanisms enable the NL2SQL agent to sustain extended refinement chains---producing accurate queries without losing critical information, while keeping latency and token cost low.

\subsubsection{Structured Context Compression}
Unbounded and unstructured context growth during iterative retry loops poses significant challenges for agents, leading to “context rot” - loss of meaningful information within the noise of an over-sized context window, which ultimately reduces response quality. To further improve context preservation beyond pruning and selective retention, we use a structured context compression approach.

After each failed SQL attempt, the agent compacts the accumulated raw conversation history into a single, structured message comprising of: (1) the original natural language question, (2) a formatted history of all prior SQL attempts annotated with their outcomes, (3) the most recent error message, and (4) an instruction to avoid repeating previously failed strategies. This structured summary ensures that the context window remains bounded, regardless of the number of retries, while preserving essential diagnostic signal for the next generation step. In parallel, schema metadata is provided in a compact DDL (Data Definition Language) format, where database tables are rendered as CREATE TABLE statements with column descriptions embedded as inline SQL comments, minimizing token overhead while preserving the semantic richness needed for accurate query generation. This bounded, information-dense representation allows the agent to incorporate failure signals across retries using fewer tokens, improving query accuracy without the quality degradation typically associated with long, noisy context windows.

\subsubsection{Agentic Tools and MCP Servers}

LLM agents, although they are very good at solving generic planning and orchestration, often fall short in executing precisely targeted actions (such as complex mathematical operations) as well as direct implementations of those actions as functions \citep{schick2023toolformer}. These actions can be writing code, scheduling meetings, or interacting with the environment such as fetching knowledge from knowledge bases. This is typically solved by implementing tools that the agents are provided with to perform that specific action. For SQL generation, function calling has been shown to help improve predictions by first converting the NL question into a structured format, before sending it to execution engine \citep{shorten2025}.

\begin{figure}[t]
\centering
\includegraphics[width=0.9\textwidth]{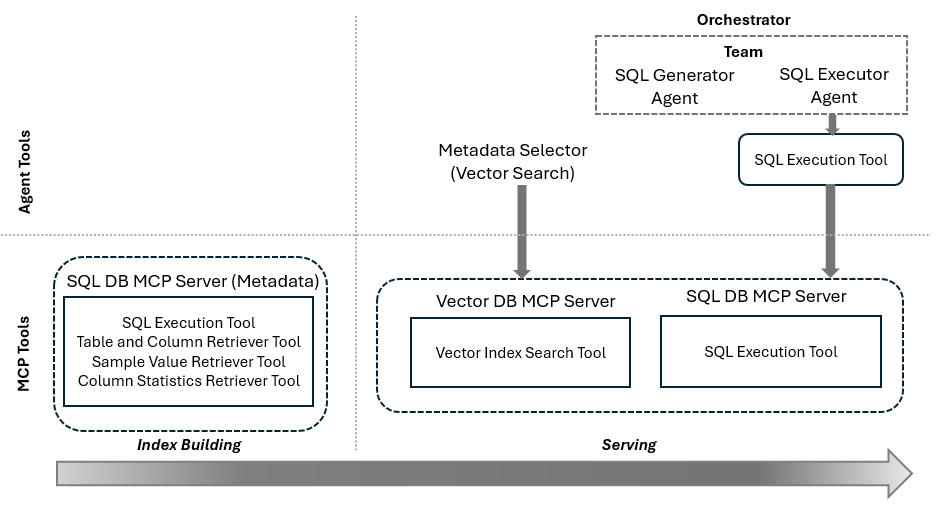}
\caption{MCP Server and agentic tools setup: Agentic tools perform higher level functions, and MCP tools perform atomic functions specific to the user's use case. Agentic tools have custom, higher level logic that has separation of concerns from the MCP atomic functions.}
\label{fig:mcp}
\end{figure}

To effectively use abstractions of such functions, we use Model Context Protocol \citep{anthropic2024mcp}. We provide agents with MCP servers with tools defined to interact with the vector index and the SQL database. MCP tools define lower-level functions that perform atomic functions such as calling SQL or knowledge base APIs. Agent tools outside of MCP, on the other hand, assist the agents to perform higher level tasks, such as searching for a keyword, and executing queries with customized inputs and/or additional logic (which in turn may invoke an MCP tool). (Figure~\ref{fig:mcp})


\begin{table}[ht]
\centering
\small
\renewcommand{\arraystretch}{1.25}
\setlength{\tabcolsep}{8pt}
\begin{tabular*}{\textwidth}{@{\extracolsep{\fill}}p{2.5cm}p{2.5cm}p{9cm}@{}}
\toprule
\textbf{MCP Server} & \textbf{Tool} & \textbf{Task} \\
\midrule
Vector DB MCP & FAISS Search & Performs vector similarity search over embedded schema descriptions and historical queries to retrieve relevant metadata. \\
SQL DB MCP & SQL Execution & Executes generated SQL queries with configurable row limits for intermediate result validation and\nobreakspace final answer retrieval. \\
\bottomrule
\end{tabular*}
\vspace{0.5em}
\caption{Agentic Tools that use MCP Servers.}
\label{tab:mcp_tools}
\end{table}

\subsection{Guardrails}

Guardrails are an important part of safety of the solution. While LLM API providers offer general guardrails (e.g., content safety), thematic agents such as NL2SQL require additional use-case-specific guardrails. The primary goal for NL2SQL is to generate SQL queries, and the agents should not modify/change the database in any manner. Hence, we implemented functional guardrails that prevent DML (Data Manipulation Language) operations that modify the database. These guardrails prevent insert, update, delete, merge, create, alter, drop, truncate, replace statements from executing, in the unlikely scenario the SQL generator generates them.

\section{Results}

\subsection{Datasets}

To evaluate the generalizability and robustness of the multi-agent NL2SQL approach, we test on the popular BIRD-SQL benchmark \citep{birdsql2025}. This benchmark, with its large-scale schema and cross-table dependencies over a million rows and multiple domains (such as finance, healthcare, sports), provides a rigorous testbed for evaluating both the accuracy and the generalizability of text-to-SQL models. For this study, we use the dev set. We initially use the financial subset to do ablation studies. After that, we test on the remaining 10 domains (which were held out and are unseen during ablations).

We use two subsets of BIRD: the BIRD 11-domain (Full Dev Set), which enables assessment of model generalizability across diverse schema structures, and the BIRD Financial Dev Set, which focuses on financial queries. Table~\ref{tab:datasets} summarizes the key characteristics of each dataset.

\begin{table}[ht]
\centering
\small
\renewcommand{\arraystretch}{1.25}
\setlength{\tabcolsep}{4pt}
\begin{tabular}{lccccl}
\toprule
\textbf{Dataset} & \textbf{Columns} & \textbf{Tables} & \textbf{DB Rows} & \textbf{Questions} & \textbf{Evaluation Focus} \\
\midrule
BIRD Financial & 55 & 8 & 1,079,680 & 106 & Financial domain knowledge and SQL generation \\
BIRD 11-domain (Full Dev Set) & 806 & 79 & 3,932,759 & 1534 & Generalizability to non-finance domains \\
\bottomrule
\end{tabular}
\vspace{0.5em}
\caption{Key characteristics of each test dataset, including schema size, number of tables, total rows, number of test questions, and evaluation focus.}
\label{tab:datasets}
\end{table}

\subsection{Accuracy Metric}

We evaluate SQL generation quality using a human-aligned LLM-based judge that assigns detailed result codes to model outputs, as described in Table~\ref{tab:error_codes}. We use GPT-4o as the LLM judge (see Appendix~\ref{app:eval_prompt} for the full evaluation prompt) and validated the LLM judge against human evaluations to ensure alignment. Accuracy is then computed based on the error codes classified by the LLM judge:

\begin{equation}
\text{Accuracy} = \frac{\text{Number of RES3 and RES5 samples}}{\text{Total evaluated cases}} \times 100\%
\end{equation}

where ``RES'' indicates result codes. Total cases are evaluated under criteria RES1--RES6. The LLM judge demonstrated over 97\% agreement (as estimated by accuracy, precision, recall, and F1-score) with human evaluation. Details of human alignment are provided in Appendix~\ref{app:human_alignment} (Table~\ref{tab:human_alignment} and Table~\ref{tab:confusion_matrices}).

\begin{table}[ht]
\centering
\small
\renewcommand{\arraystretch}{1.3}
\setlength{\tabcolsep}{10pt}
\begin{tabular}{lp{8cm}c}
\toprule
\textbf{Code} & \textbf{Description} & \textbf{Example (Ground Truth vs. Generated)}\\
\midrule
RES1 & Failed Execution: Query failed due to errors or missing tables. & [(42,)] vs. Error\\
RES2 & Executed, Incorrect Result: Query ran successfully, but the output does not contain the ground truth. & [(42,)] vs. (99,)\\
RES3 & Executed, Correct Result: The output matches ground truth exactly, with no extraneous data. Aligning with human evaluation. & [(42,)] vs. (42,) \\
RES4 & Executed, No Result: Query ran but returned no data, when a result was expected. & [(42,)] vs. [] \\
RES5 & Executed, Partial Match: Query output includes ground truth along with other values. & [(42,)] vs. (42,99)*\\
RES6 & Unexpected Result: Query returned a malformed or unintended result. & [(42,)] vs. "Failed to execute the query" \\
\bottomrule
\end{tabular}
\vspace{0.5em}
\caption{Criteria Definitions (Result Code Descriptions).$^\ast$ In practice, query outputs are returned as pandas DataFrames with column headers as well, providing context for interpreting the values (e.g., \texttt{[\{"account\_id": 42, "client\_id": 99\}]}).}
\label{tab:error_codes}
\end{table}

\subsection{Results \& Discussion}

\subsubsection{Accuracy}

Our experiments spanned a variety of model configurations, ranging from baseline non-orchestrated agents to advanced hybrid ensembles, and included systematic ablation studies to isolate the contributions of individual system components. Baseline models, such as single-agent configurations using GPT-4o and prompt engineering variants, provided a baseline metric for NL2SQL accuracy without advanced schema enrichment or embedding-based vector search for relevant column retrieval. This setting yielded an accuracy of 60.2\%. This initial performance underscores the inherent challenges of NL2SQL, where queries often require complex joins, precise schema understanding, nuanced handling of domain-specific logic, and prompt engineering to manage long contexts. Table~\ref{tab:ablation} lists all the successive improvements over the baseline.

For the baseline accuracy result discussed in Table~\ref{tab:ablation}, we used a simple Selector Group Chat, akin to prior research in Actor-Critic approaches for multi-agent LLM collaboration \citep{estornell2024acc}. Since the selector agent flow is not fully dynamic (typically dependent on manually configured agentic workflow cycles), we improved upon it by adding smart orchestration mechanisms using Autogen and vector search capabilities. The orchestration component coordinates multiple agents, enabling more effective division of labor and context management, while vector search facilitates efficient retrieval of relevant schema elements and metadata. The integration of these components resulted in an accuracy improvement rising to the 68--72\% range (Table~\ref{tab:ablation}). These gains were particularly evident in queries involving intricate table relationships and long join paths, which are common in financial datasets.

Further enhancements were achieved through multi-model agents. In this configuration, each agent is assigned the LLM best suited to its specialized task. For planning and reasoning, OpenAI GPT models are known to perform well, whereas for code generation, Anthropic's Claude models have topped recent leaderboards. By combining the strengths of multiple LLMs, we increased accuracy to 76.4\% on the BIRD financial dataset (Table~\ref{tab:ablation}). These results were obtained using the Autogen orchestrator, which is not efficient in terms of managing long contexts, resulting in higher latency. 

For additional improvements, we developed our custom orchestrator along with structured context compression. Together, these enhancements yielded the highest accuracy of 78.1 \% using Opus 4.1 as the SQL generator. Table~\ref{tab:accuracy_domains} further demonstrates the improvements achieved by introducing our new NL2SQL orchestrator across diverse domains. 

To establish an upper bound for model performance, we included human expert evaluation sourced from the BIRD bench leaderboard. Human experts, with deep domain knowledge and experience in SQL query formulation, achieved an accuracy of 93\% on entire BIRD test dataset, serving as a directional indicator of expert-level SQL-writing accuracy. We ran ablation studies to quantify the impact of individual system components, as shown in Table~\ref{tab:ablation}, demonstrating that systematic integration of orchestration, retrieval, and advanced model selection leads to substantial improvements in NL2SQL task performance. 


\begin{table}[ht]
\centering
\small
\begin{tabular}{lcc}
\toprule
\textbf{Improvements} & \textbf{Accuracy (\%)} \\
\midrule
Baseline (Pure sequential, config-based agent flow) & 60.2 \\
+ Dynamic agent orchestration with planner & 66.7 \\
+ Vector search (1-shot) & 68.5 \\
+ Enriched Schema (0-shot) & 70.8 \\
+ Prompt Engineering (0-shot) & 71.6 \\
+ 1-shot & 72.7 \\
+ Opus 4.1 as SQL generator (0-shot) & 76.4 \\
+ Structured Context Compression & 79.2 \\ 
\bottomrule
\end{tabular}
\vspace{0.5em}
\caption{Incremental accuracy gains achieved through successive enhancements to the agent architecture for NL2SQL tasks on the BIRD-bench financial dataset.}
\label{tab:ablation}
\end{table}

\subsubsection{Generalization Across Domains}

To examine the generalizability of our approach to different domains beyond the financial domain, we conducted a comparative evaluation for BIRD's 11 domains with different configurations, as described in Table~\ref{tab:configurations}.

\begin{table}[ht]
\centering
\small
\renewcommand{\arraystretch}{1.25}
\setlength{\tabcolsep}{8pt}
\begin{tabular}{ll}
\toprule
\textbf{Version} & \textbf{Configuration} \\
\midrule
V1 & GPT-4o, without vector search, one-shot \\
V2 & GPT-4o, vector search, enriched metadata, zero-shot \\
V3 & Hybrid model GPT-4o/Opus 4.1 (V2 with Query writing agent now using Opus 4.1) \\
V4 & V2 with our enhanced new custom orchestrator \\
V5 & V3 with our enhanced new custom orchestrator \\
V6 & V5 with structured prompt (context management) \\
\bottomrule
\end{tabular}
\vspace{0.5em}
\caption{Description of different configurations used for evaluation.}
\label{tab:configurations}
\end{table}



Focusing first on the Autogen orchestrator section of Table~\ref{tab:accuracy_domains}, V3 -- which uses Opus 4.1 as the SQL generator agent and GPT-4o for all other agents (Figure~\ref{fig:architecture}) -- achieved the highest accuracy at 75.4 \%, outperforming baseline models in nearly every domain. V2, which uses GPT-4o for all agents with zero-shot vector search, attained an average accuracy of 72.5 percent, higher than V1 (68.5\%). Notably, V2's performance demonstrates that integrating efficient vector search enables the model to match or exceed the accuracy of V1 without requiring one-shot examples, thereby simplifying the inference process. V3 not only delivered the best overall results but also exhibited greater consistency across domains, highlighting the impact of advanced model architectures on generalization and robustness. The largest accuracy gains were observed in schema-dense domains, where database complexity demands sophisticated reasoning and precise schema understanding. These findings suggest that both architectural improvements and retrieval-augmented methods contribute to enhanced NL2SQL performance, with vector search offering a practical alternative to manual example selection.

Next, we applied our custom orchestrator to scenarios V4, V5, and V6. The new design, with its optimized planning and progress self-revisions, along with the optimized inference prompt, yields additional accuracy improvement in aggregate across all domains. For both non-hybrid and hybrid setups, these optimizations improved the accuracy from 72.5\% to 76.0\% (3.5\% gain for non-hybrid), and 75.4\% to 76.3\% (0.9\% gain for hybrid setup). Structured context compression yielded an additional 1.8\% absolute improvement, resulting in an overall accuracy of 78.1\% on the 11-domain BIRD benchmark. In addition, it reduced token usage by approximately 19.8\% on the BIRD-bench financial dataset.

\begin{table}[t]
\centering
\small
\renewcommand{\arraystretch}{1.25}
\setlength{\tabcolsep}{5pt}
\begin{tabular}{l*{6}{c}}
\toprule
 & \multicolumn{3}{c}{\textbf{Autogen Orchestrator}} & \multicolumn{3}{c}{\textbf{Custom Orchestrator}} \\
\cmidrule(lr){2-4} \cmidrule(lr){5-7}
\textbf{Domain} & \makecell{V1\\{\scriptsize(No Vec.}\\\scriptsize{Search)}} & \makecell{V2\\{\scriptsize(+Vector}\\\scriptsize{Search)}} & \makecell{V3\\{\scriptsize(+Hybrid}\\\scriptsize{Models)}} & \makecell{V4\\{\scriptsize(+Vector}\\\scriptsize{Search)}} & \makecell{V5\\{\scriptsize(+Hybrid}\\\scriptsize{Models)}} & \makecell{V6\\{\scriptsize(+Struct.}\\\scriptsize{Context)}} \\
\midrule
California Schools & 55.7 & 67.4 & 73.0 & 74.2 & 78.7 & 75.3 \\
Codebase Community & 61.1 & 63.3 & 70.4 & 73.1 & 73.7 & 76.8 \\
Debit Card Specializing & 59.4 & 76.5 & 79.6 & 64.2 & 65.6* & 71.9 \\
European Football 2 & 72.9 & 75.0 & 80.6 & 78.3 & 75.8 & 79.1 \\
Financial & 66.7 & 70.8 & 76.4 & 75.5 & 76.4 & 79.2 \\
Formula 1 & 69.5 & 69.4 & 71.2 & 70.1 & 71.8 & 70.7 \\
Student Club & 83.5 & 83.5 & 87.9 & 88.0 & 86.7 & 91.1 \\
Superhero & 88.4 & 89.2 & 89.9 & 94.6 & 95.3 & 96.1 \\
Toxicology & 67.6 & 81.4 & 81.4 & 83.4 & 84.1 & 84.1 \\
Thrombosis Prediction & 60.1 & 64.4 & 60.7 & 62.6 & 62.6 & 63.2 \\
Card Games & 64.2 & 64.5 & 67.5 & 71.2 & 69.6 & 73.3 \\
\midrule
\textbf{Weighted Average} & \textbf{68.5} & \textbf{72.5} & \textbf{75.4} & \textbf{76.0} & \textbf{76.3} & \textbf{78.1} \\
\bottomrule
\end{tabular}
\vspace{0.5em}
\caption{NL2SQL Accuracy (\%) of LLM Configurations Across Diverse Tasks. The scenarios V1--V6 are listed in Table~\ref{tab:configurations}. All ablations were done using financial dataset only, and the remaining 10 domains were held out to assess generalizability. $^\ast$Debit Card Specialization dataset has inconsistencies in the ground truth hence V5 accuracy is underestimated. See Table~\ref{tab:inconsistencies} in Appendix~\ref{app:inconsistencies} for examples.}
\label{tab:accuracy_domains}
\end{table}

\subsubsection{Confidence Intervals}

To estimate confidence intervals, we carried out 5 trials each with each of the models used for query generation, on financial data.

\begin{figure}[ht]
\centering
\includegraphics[width=0.8\textwidth]{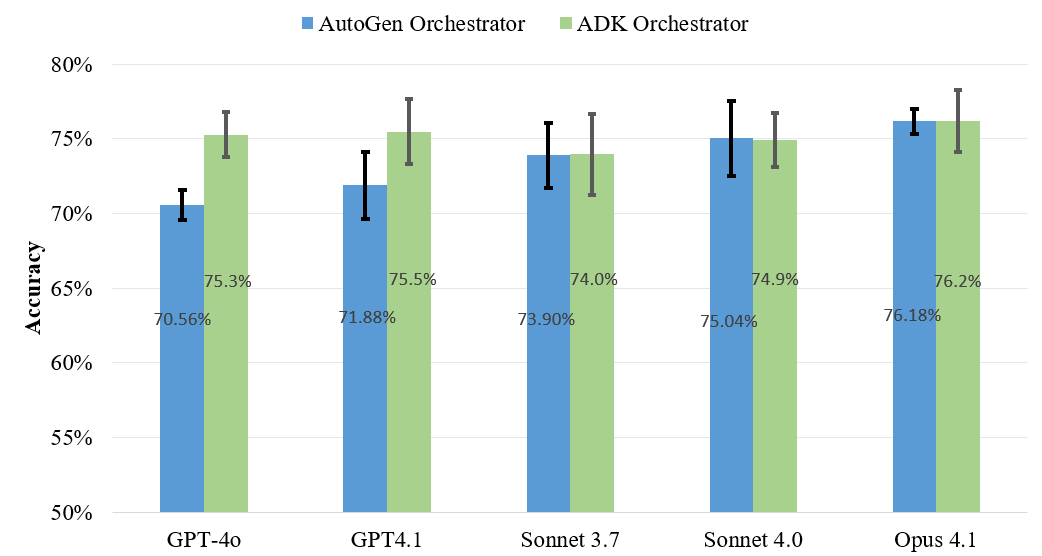}
\caption{Accuracy on BIRD financial dataset with different SQL generator LLMs. Orchestrator in all cases used GPT-4o as the LLM. Confidence intervals indicate 95\% margin of error with $t$-distribution. See details in Appendix~\ref{app:stat_significance} for individual trials.}
\label{fig:confidence}
\end{figure}

Due to the computational cost of computing confidence intervals across all possible configurations, we restrict our analysis to the BIRD financial dataset, comprising 106 test samples. The SQL query generator agent uses GPT-4o with a temperature of 0.3, $\mathrm{top\_p} = 1$. Given the small number of trials (5), we used the 2-sided $t$-distribution to compute standard errors and confidence intervals at 95\% significance. The results shown in Figure~\ref{fig:confidence} indicate that Opus 4.1 outperforms the other models, achieving the highest average accuracy of 76.18\% and demonstrating the greatest consistency with the lowest margin of error ($\pm 0.83$). Sonnet 4.0 follows closely with an average accuracy of 75.04\%, while Sonnet 3.7 also performs well at 73.90\%. The GPT models, GPT-4.1 and GPT-4o, lag with average accuracies of 71.88\% and 70.56\%, respectively. (See Table~\ref{tab:stat_significance} in Appendix~\ref{app:stat_significance} for accuracy on individual trials.)

\subsubsection{Latency}

\begin{figure}[ht]
\centering
\begin{subfigure}[b]{0.48\textwidth}
\centering
\includegraphics[width=\textwidth]{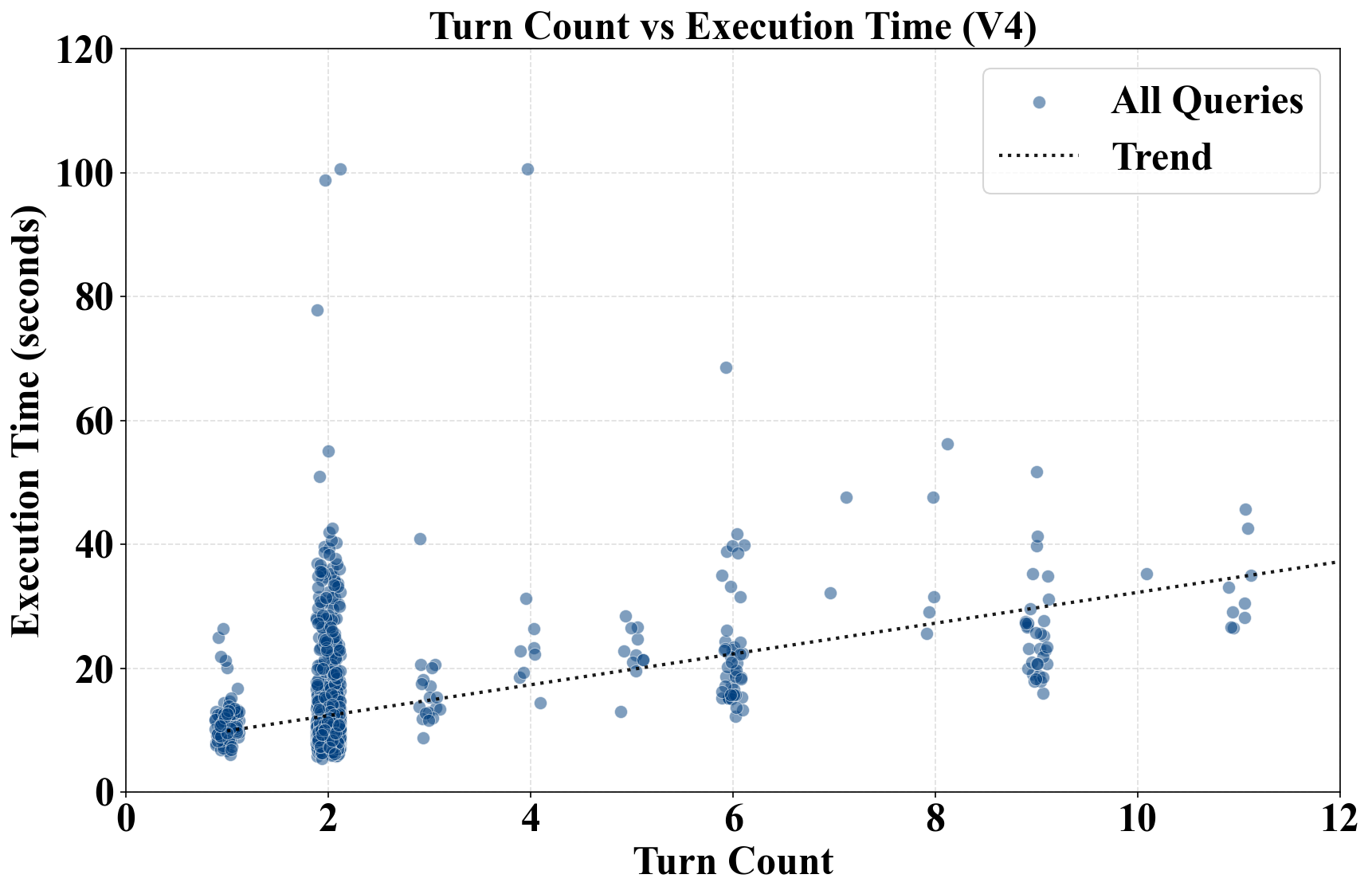}
\caption{GPT-4o as SQL writer model}
\label{fig:latency_gpt}
\end{subfigure}
\hfill
\begin{subfigure}[b]{0.48\textwidth}
\centering
\includegraphics[width=\textwidth]{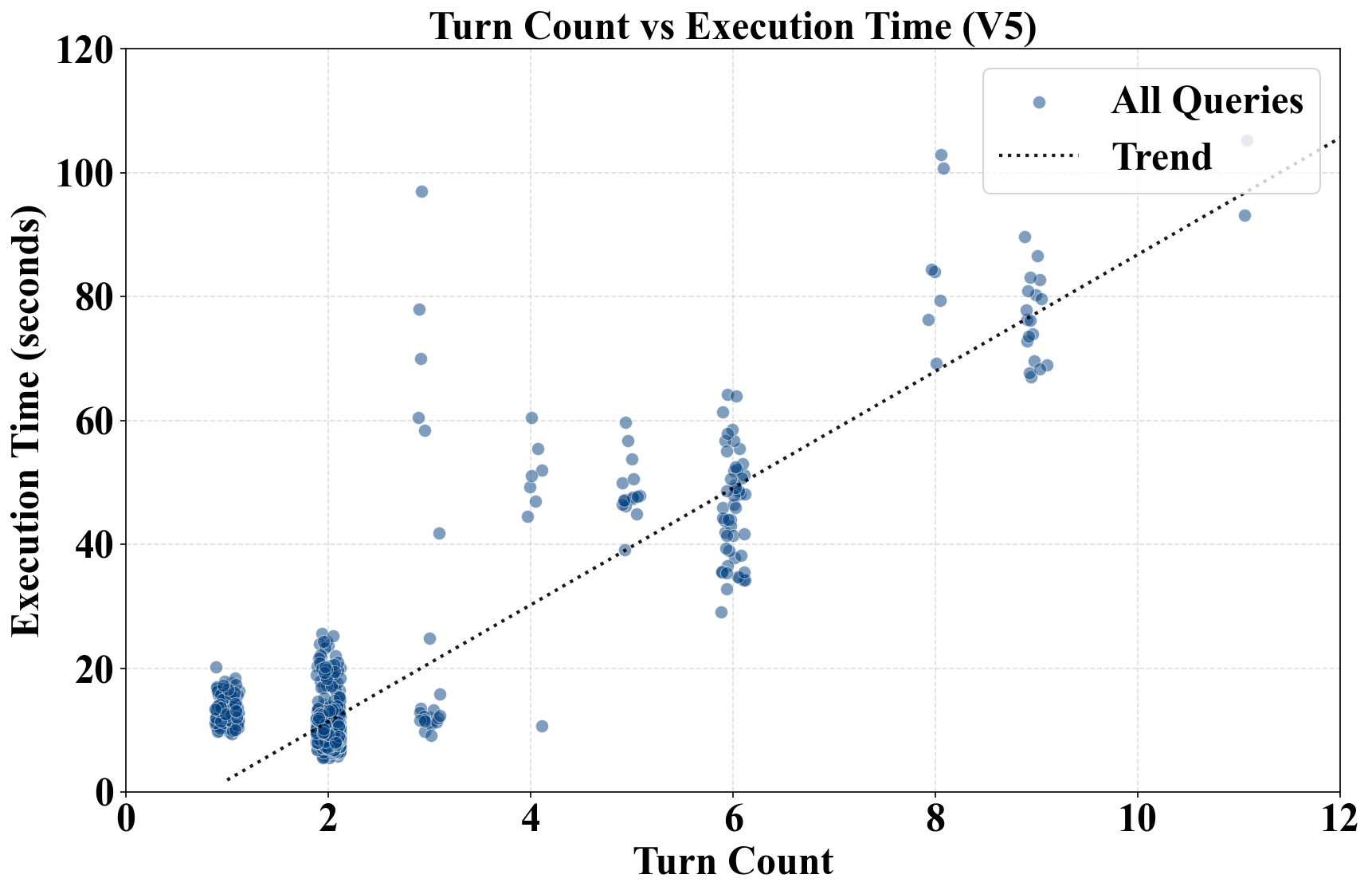}
\caption{Claude Opus 4.1 as SQL writer model}
\label{fig:latency_opus}
\end{subfigure}
\caption{Correlation between latency and number of turns of thematic NL2SQL agent. Number of turns is defined as number of invocations of SQL writer agent and executor agent. Larger execution time is caused by exceptionally complex queries requiring multiple self-reflective loops.}
\label{fig:latency}
\end{figure}

For complex queries, agents may require more iterations to replan and adjust, resulting in longer execution times. We verified this hypothesis by plotting the number of turns required against execution time, which reveals a clear correlation (Figure~\ref{fig:latency}). Notably, for some complex queries at the tail end of the distribution, the agent requires significantly more turns to arrive at the correct query. Latency on Bedrock models is generally higher, hence the overall latency for the Claude option in Table~\ref{tab:latency} is higher. We see occasional outliers where execution times were high due to API issues or some rare, failed samples; such failures are penalized when computing the accuracy score.

\begin{table}[ht]
\centering
\small
\begin{tabular}{llccc}
\toprule
\textbf{Model} & \textbf{Version} & \textbf{P50 Latency (s)} & \textbf{P90 Latency (s)} & \textbf{P95 Latency (s)} \\
\midrule
Claude Opus 4.1 & V5 (New Orchestrator) & 10.3 & 17.8 & 44.8 \\
Claude Opus 4.1 & V3 & 134.9 & 168.6 & 192.9 \\
GPT 4o & V4 (New Orchestrator) & 11.3 & 18.6 & 26.5 \\
GPT 4o & V2 & 58.7 & 89.2 & 93.1 \\
\bottomrule
\end{tabular}
\vspace{0.5em}
\caption{Latency of the NL2SQL thematic agent for different model configurations on BIRD financial dataset. Both LLMs give similar latency, and the new orchestrator achieves substantial improvement in latency.}
\label{tab:latency}
\end{table}

\section{Conclusion and Future Work}

We built a NL2SQL system for enterprise use that uses a multi-agent orchestration over enriched schema metadata with retrieval, planning, reflection, and guarded execution. The approach handles large, complex schemas by finding the most relevant tables and columns for a given query. Across the 11-domain subsets of the BIRD benchmark dataset, the system consistently outperforms non-agentic approach. Vector search and schema enrichment raise semantic accuracy more than exact match and reduce number of query iterations resulting in low latency. The agent is able to attain over 78.1\% semantic accuracy on BIRD dataset. Further improvement avenues include extending the agent to handle various SQL dialects and domains, and enhancing search and ranking for more accurate retrieval.

\section{Limitations}

Given the limitations of LLMs in producing precise code, which is still an area of research, we recommend a human in the loop for critical applications. LLM pretraining data (depending on the model used) may not always have a lot of SQL data, hence the model may sometimes struggle to produce extremely complex queries correctly. Furthermore, coverage of different SQL dialects and database schemas with specialized domains may be limited. Therefore, user as an SME is recommended to provide business rules/knowledge in the context, for which we provide input parameters. Regarding database size, extremely large databases (hundreds or tables, thousands of columns, and hundreds of millions of rows) may require additional accuracy validation, enhancements for retrieval, and latency optimizations.

\subsubsection*{Acknowledgements}
We thank Dawid Bernaciak, Ashish Singla, Ashish Garg, Leo Carrera, Dipesh Singh, Sagar Dashora and other members of the JPMC agents team for their collaboration.

\bibliographystyle{plainnat}
\bibliography{references}
\begin{appendices}

\section{Evaluation Prompt}
\label{app:eval_prompt}

The full evaluation prompt used by the Eval agent is below.\footnote{Both the ground truth and generated tables are truncated to the first 500 characters to avoid context length errors caused by excessively long outputs. Testing confirmed that this truncation threshold has negligible effect on results---truncating at 500 versus 10,000 characters yielded only a \textasciitilde0.3\% difference in overall accuracy.}

\begin{tcolorbox}[colback=block-gray,grow to right by=0mm,grow to left by=0mm,boxrule=0pt,boxsep=0pt,breakable]
Classify each output from the SQL Execution Agent according to the following criteria and the ground truth and the generated answer provided to you:\\

\phantom{00}The ground truth answer is: \{ground\_truth\_table\}\\
\phantom{00}The generated answer is: \{generated\_table\}\\

The classification codes are:\\
\phantom{00}RES1. Failed Execution: Query failed due to syntax errors, missing tables, or other runtime issues.\\
\phantom{00}RES2. Executed, Incorrect Result: Query executed but the result does not match the ground truth, and the ground truth is NOT present in the answer.\\
\phantom{00}RES3. Executed, Correct Result: Query executed and the result matches the ground truth (only cases with brackets or syntax -- NOT other numbers or letters).\\
\phantom{00}RES4. Executed, No Result: Query executed but returned no rows or \texttt{None}.\\
\phantom{00}RES5. Executed, Partial Match: Query executed and the result partially matches the ground truth, where ground truth exists but there ARE other numbers or letters.\\
\phantom{00}RES6. Unexpected Result: Query executed but returned a malformed or unexpected result.\\

Examples:\\
\phantom{00}RES3: ground truth: [(71,)] $\rightarrow$ generated answer: (71,)\\
\phantom{00}RES3: ground truth: [(71, Bank)] $\rightarrow$ generated answer: (Bank, 71)\\
\phantom{00}RES5: ground truth: [(71,)] $\rightarrow$ generated answer: (71, 72)\\
\phantom{00}RES4: ground truth: [(71,)] $\rightarrow$ generated answer: None\\
\phantom{00}RES2: ground truth: [(71,)] $\rightarrow$ generated answer: (72,)\\

RETURN a JSON:\\
\texttt{\{\{"Classification code": "<add it here>", "Reasoning": "<explanation>"\}\}}
\end{tcolorbox}

\section{Human alignment of the LLM judge}
\label{app:human_alignment}
We compared human labeled ground truth (RES error codes) and LLM judge generated labels to evaluate the human alignment of our LLM judge. To avoid human bias, two different human judges on our team independently evaluated the LLM judge's results on two separate datasets. We selected all 106 BIRD financial dev data samples and all 129 BIRD European Football subset samples for this comparison and marked instances where the LLM judge did not align (agree) with the human judge.

\begin{table}[ht]
\centering
\small
\begin{tabular}{lcc}
\toprule
\textbf{Metric (\%)} & \textbf{BIRD-Financial} & \textbf{BIRD - European Football} \\
\midrule
Accuracy (Alignment Rate) & 98.11 & 98.45 \\
Precision & 97.3 & 97.7 \\
Recall & 100 & 100 \\
F1-Score & 98.63 & 98.84 \\
\bottomrule
\end{tabular}
\vspace{0.5em}
\caption{Results of human alignment assessment. With the evaluation judge prompt, we get 97\% or above alignment between human and LLM judge, for all metrics (precision, recall, F1-score, accuracy).}
\label{tab:human_alignment}
\end{table}


\begin{table}[ht]
\centering
\small
\renewcommand{\arraystretch}{1.25}
\setlength{\tabcolsep}{6pt}
\begin{minipage}[t]{0.48\textwidth}
\centering
\textbf{BIRD Financial}\\[0.5em]
\begin{tabular}{lcc}
\toprule
 & \makecell{Human label\\= Correct} & \makecell{Human label\\= Incorrect} \\
\midrule
\makecell[l]{LLM Judge Label\\= Correct} & \makecell{72\\(True Positives)} & \makecell{2\\(False Positives)} \\
\makecell[l]{LLM Judge Label\\= Incorrect} & \makecell{0\\(False Negatives)} & \makecell{32\\(True Negatives)} \\
\bottomrule
\end{tabular}
\end{minipage}%
\hfill
\begin{minipage}[t]{0.48\textwidth}
\centering
\textbf{BIRD European Football}\\[0.5em]
\begin{tabular}{lcc}
\toprule
 & \makecell{Human label\\= Correct} & \makecell{Human label\\= Incorrect} \\
\midrule
\makecell[l]{LLM Judge Label\\= Correct} & \makecell{85\\(True Positives)} & \makecell{2\\(False Positives)} \\
\makecell[l]{LLM Judge Label\\= Incorrect} & \makecell{0\\(False Negatives)} & \makecell{42\\(True Negatives)} \\
\bottomrule
\end{tabular}
\end{minipage}
\vspace{0.5em}
\caption{Confusion matrices for human alignment assessment.}
\label{tab:confusion_matrices}
\end{table}

\section{Statistical significance}
\label{app:stat_significance}

\noindent
\begin{minipage}{\textwidth}
\centering
\small
\renewcommand{\arraystretch}{1.25}
\setlength{\tabcolsep}{8pt}
\begin{tabular}{lccccc}
\toprule
\textbf{Trial} & \textbf{GPT-4o} & \textbf{GPT-4.1} & \textbf{Sonnet 3.7*} & \textbf{Sonnet 4.0*} & \textbf{Opus 4.1*} \\
\midrule
Trial 1 & 75.5 & 73.6 & 71.7 & 74.5 & 73.6 \\
Trial 2 & 74.5 & 77.4 & 72.6 & 74.5 & 76.4 \\
Trial 3 & 76.4 & 73.6 & 73.6 & 77.4 & 78.3 \\
Trial 4 & 73.6 & 76.4 & 77.4 & 73.6 & 76.4 \\
Trial 5 & 76.4 & 76.4 & 74.5 & 74.5 & 76.4 \\
\midrule
Average & 75.3 & 75.5 & 74.0 & 74.9 & 76.2 \\
Margin of Error ($\pm$) & 1.52 & 2.19 & 2.72 & 1.80 & 2.09 \\
\bottomrule
\end{tabular}
\vspace{0.5em}
\captionof{table}{Confidence Interval and Accuracy Comparison of Thematic NL2SQL multi-agent solution. *Multi-model setup where SQL query generator is Anthropic model and other agents as discussed in Figure~\ref{fig:architecture} use GPT-4o.}
\label{tab:stat_significance}
\end{minipage}

\section{Inconsistencies in ground truth}
\label{app:inconsistencies}

BIRD SQL benchmark dataset is known to have inconsistencies in the ground truth, as examined by \citet{wretblad2024}. To compare fairly with public leaderboard, we do not adjust any ground truth in this paper. But we notice that in our results with the V5 variant (which is our best variant), the ``debit card specializing'' domain shows lower accuracy than prior iterations. Hence we examined this domain in more details for ground truth validity.

Below examples show that in some of the examples, the agent's answer is correct and the ground truth is incorrect. In these cases, the ground truth ignores the definition of ``total price = Amount $\times$ Price'' provided in the original benchmark data, and generates queries where it incorrectly calculates ``Revenue'' and ``Average Total Price'' using price, while not accounting for volume. After correcting for it, our agent's accuracy increases from 65.6 to 71.9.

\begin{table}[ht]
\begin{center}
\small
\resizebox{\textwidth}{!}{%
\begin{tabular}{p{3.5cm}p{5.5cm}p{5.5cm}}
\toprule
\textbf{Question} & \textbf{Ground Truth SQL} & \textbf{Agent SQL} \\
\midrule
What is the average total price of the transactions taken place in gas stations in the Czech Republic?
& \texttt{SELECT AVG(T1.Price) FROM transactions\_1k AS T1 INNER JOIN gasstations AS T2 ON T1.GasStationID = T2.GasStationID WHERE T2.Country = 'CZE'}
& \texttt{SELECT AVG("transactions\_1k"."Amount" * "transactions\_1k"."Price") AS "average\_total\_price" FROM "transactions\_1k" JOIN "gasstations" ON "transactions\_1k"."GasStationID" = "gasstations"."GasStationID" WHERE "gasstations"."Country" = 'CZE'} \\
\midrule
What is the average total price of the transactions taken place in January, 2012?
& \texttt{SELECT AVG(Amount) FROM transactions\_1k WHERE Date LIKE '2012-01\%'}
& \texttt{SELECT AVG("Amount" * "Price") AS "average\_total\_price" FROM "transactions\_1k" WHERE "Date" LIKE '2012-01\%'} \\
\midrule
For the customers who paid in the euro, what is their average total price of the transactions?
& \texttt{SELECT AVG(Amount) FROM transactions\_1k WHERE Date LIKE '2012-01\%'}
& \texttt{SELECT AVG("transactions\_1k"."Amount" * "transactions\_1k"."Price") AS "average\_total\_price" FROM "customers" INNER JOIN "transactions\_1k" ON "customers"."CustomerID" = "transactions\_1k"."CustomerID" WHERE "customers"."Currency" = 'EUR'} \\
\midrule
Which gas station has the highest amount of revenue?
& \texttt{SELECT GasStationID FROM transactions\_1k GROUP BY GasStationID ORDER BY SUM(Price) DESC LIMIT 1}
& \texttt{SELECT "gasstations"."GasStationID", SUM("transactions\_1k"."Amount" * "transactions\_1k"."Price") AS "total\_revenue" FROM "transactions\_1k" JOIN "gasstations" ON "transactions\_1k"."GasStationID" = "gasstations"."GasStationID" GROUP BY "transactions\_1k"."GasStationID" ORDER BY "total\_revenue" DESC LIMIT 1} \\
\midrule
What is the amount spent by customer ``38508'' at the gas stations? How much had the customer spent in January 2012?
& \texttt{SELECT SUM(T1.Price), SUM(IIF(T3.Date = '201201', T1.Price, 0)) FROM transactions\_1k AS T1 INNER JOIN gasstations AS T2 ON T1.GasStationID = T2.GasStationID INNER JOIN yearmonth AS T3 ON T1.CustomerID = T3.CustomerID WHERE T1.CustomerID = '38508'}
& \texttt{SELECT SUM("transactions\_1k"."Amount" * "transactions\_1k"."Price") AS "total\_spent\_at\_gas\_stations", SUM("yearmonth"."Consumption") AS "total\_spent\_in\_january\_2012" FROM "transactions\_1k", "yearmonth" WHERE "transactions\_1k"."CustomerID" = 38508 AND "transactions\_1k"."GasStationID" IS NOT NULL AND "yearmonth"."CustomerID" = 38508 AND "yearmonth"."Date" = '201201'} \\
\bottomrule
\end{tabular}%
}
\end{center}
\vspace{0.5em}
\caption{Sample outputs for questions about ``total price'' or ``amount,'' which show ground truth ignoring the evidence (``total price = Amount $\times$ Price'') provided in the original schema, while the agent correctly uses the given business logic to generate the correct SQL query.}
\label{tab:inconsistencies}
\end{table}

\end{appendices}

\end{document}